\documentclass[letterpaper, 10 pt, conference]{ieeeconf}  

\IEEEoverridecommandlockouts                              

\usepackage{algorithm}
\usepackage{algorithmic}

\usepackage{amsmath} 
\usepackage{amsfonts} 
\usepackage{amsfonts}       

\usepackage{multirow}
\usepackage{graphicx}
\usepackage{varioref}

\usepackage{subcaption}
\usepackage{booktabs}
\usepackage{dsfont}
\usepackage{xcolor}         
\usepackage[hyphens]{url}

\DeclareMathOperator*{\argmin}{argmin}
\DeclareMathOperator*{\argmax}{argmax}

\newcommand{\ie}{\textit{i.e.}}

\title{\LARGE \bf
Mitigating Adversarial Perturbations for \\Deep Reinforcement Learning via Vector Quantization
}

\author{Tung M. Luu$^{1}$, Thanh Nguyen$^{1}$, Tee Joshua Tian Jin$^{1}$, Sungwoon Kim$^{2}$, and Chang D. Yoo$^{1*}$
    \thanks{$^*$Corresponding author: Chang D. Yoo} %
    \thanks{$^{1}$School of Electrical Engineering, Korea Advanced Institute of Science and Technology, Daejeon 34141, Republic of Korea. {\{tungluu2203,thanhnguyen,joshuateetj,cd\_yoo\}@kaist.ac.kr}} %
    \thanks{$^{2}$Department of Artificial Intelligence, Korea University,
            145 Anam-ro, Seongbuk-gu, Seoul 02841, Republic of Korea.
            {swkim01@korea.ac.kr}}%
    \thanks{This work was partly supported by Institute for Information \& communications Technology Planning \& Evaluation (IITP) grant funded by the Korea government(MSIT) (No. 2021-0-01381, Development of Causal AI through Video Understanding and Reinforcement Learning, and Its Applications to Real Environments) and partly supported by Institute of Information \& communications Technology Planning \& Evaluation (IITP) grant funded by the Korea government(MSIT) (No.2022-0-00184, Development and Study of AI Technologies to Inexpensively Conform to Evolving Policy on Ethics).} %
}

\begin{document}

\maketitle
\thispagestyle{empty}
\pagestyle{empty}

\begin{abstract}

Recent studies reveal that well-performing reinforcement learning (RL) agents in training often lack resilience against adversarial perturbations during deployment. This highlights the importance of building a robust agent before deploying it in the real world. Most prior works focus on developing robust training-based procedures to tackle this problem, including enhancing the robustness of the deep neural network component itself or adversarially training the agent on strong attacks. In this work, we instead study an input transformation-based defense for RL. Specifically, we propose using a variant of vector quantization (VQ) as a transformation for input observations, which is then used to reduce the space of adversarial attacks during testing, resulting in the transformed observations being less affected by attacks. Our method is computationally efficient and seamlessly integrates with adversarial training, further enhancing the robustness of RL agents against adversarial attacks. Through extensive experiments in multiple environments, we demonstrate that using VQ as the input transformation effectively defends against adversarial attacks on the agent's observations. 

\end{abstract}

\section{INTRODUCTION}

Modern deep reinforcement learning (RL) agents \cite{mnih2015human,fujimoto2018addressing,haarnoja2018soft} typically rely on deep neural networks (DNN) as powerful function approximators. Nevertheless, it has been discovered that even a well-trained RL agent may drastically fail under the small adversarial perturbations in the input during deployment \cite{huang2017adversarial,lin2017tactics,kos2017delving,behzadan2017vulnerability,pattanaik2018robust}, making it risky to execute on safety-critical applications such as autonomous driving \cite{you2019advanced}. Therefore, it is necessary to develop techniques to assist the RL agents in resisting adversarial attacks in input observations before deploying them into the real world.

There have been many works proposed in the literature in defending against adversarial attacks on input observations. A line of work focuses on enhancing the robustness of DNN components by enforcing properties such as invariance and smoothness via regularization schemes \cite{shen2020deep,zhang2020robust,oikarinen2021robust,yang2022rorl}, resulting in deep policy outputs that exhibit similar actions under bounded perturbations. Another line of work considers training the RL agent in an adversarial manner, where an adversary is introduced to perturb the agent's input while it interacts with an environment. Sampled trajectories under these attacks are subsequently used for training, resulting in a more resilient RL agent. In this approach, the perturbation can be induced from the policy/value function \cite{kos2017delving,behzadan2017whatever,pattanaik2018robust,liang2022efficient} or more recently, it can be generated by another RL-based adversary \cite{zhang2021robust,sun2021strongest}. While training with RL-based attackers can attain high long-term rewards under attacks, it often requires extra samples and computations for training.

Aforementioned strategies can be regarded as robust training-based defenses, aimed at learning resilient policy networks against adversarial attacks. Meanwhile, in the field of image classification, there are also numerous input transformation-based defenses \cite{dziugaite2016study,liao2018defense,guo2018countering,prakash2018deflecting,gupta2019ciidefence,jin2019ape,salman2020denoised} that mitigate such attacks without altering the underlying model. These defenses attempt to reduce adversarial perturbations in the input by transforming it before feeding to the model. The transformation process commonly involves denoisers for purifying perturbations \cite{gupta2019ciidefence,jin2019ape,salman2020denoised,nie2022diffusion} or simply utilizes image processing techniques to weaken the effect of attacks \cite{dziugaite2016study,lu2017no,guo2018countering,xu2018feature,prakash2018deflecting}. Therefore, this approach potentially benefits RL agents without requiring significant changes to underlying RL algorithms. However, denoiser-based transformations often leverage powerful generative models such as GAN \cite{samangouei2018defense,jin2019ape} or diffusion model \cite{nie2022diffusion} to remove noise, which may introduce overhead in both training and inference for RL agents. On the other hand, the processing-based transformations are appealing due to their non-differential nature, making it challenging for adversaries to circumvent the defenses. Additionally, these transforms are also cost-efficient and versatile, making them suitable for use with RL agents. Nonetheless, many of these transformations are tailored to image data  \cite{dziugaite2016study,guo2018countering,xu2018feature,prakash2018deflecting} and may not easily extend to vector inputs such as low-dimensional states in continuous control tasks. 

Motivated by this limitation, we propose using a variant of vector quantization (VQ) as a suitable input transformation-based defense for RL, which is generally applicable for both image input and continuous state. The key idea of our approach is to utilize VQ for discretizing the observation space and subsequently train the RL agent within this transformed space. This strategy effectively reduces the space of adversarial attacks \cite{huang2017adversarial,kos2017delving,zhang2020robust} that can impact the agent's observations, producing transformed inputs that are minimally affected by attacks. Our proposed approach is computationally efficient and modifies only the input rather than the model itself, allowing it to synergistically complement other robust training-based defenses and enhance the overall robustness of the RL agent.

The main contributions of the paper are as follows: (i) we propose a novel input transformation-based defense for RL agent using VQ, (ii) we introduce an effective way to incorporate the proposed defense in the RL algorithms, and (iii) we demonstrate through extensive experiments that our proposed method effectively mitigates adversarial attacks on many environments across domains and settings.  The code can be found at https://github.com/tunglm2203/vq\_robust\_rl.

\section{RELATED WORK}\label{sec:related_work}

\textbf{Adversarial Attacks on State Observations.} Since the discovery of adversarial examples in the classification \cite{szegedy2013intriguing}, vulnerabilities in state observations of deep RL were first demonstrated by \cite{huang2017adversarial,lin2017tactics,kos2017delving}. \cite{huang2017adversarial} evaluated the robustness of DQN agents in the Atari domain using FGSM \cite{goodfellow2014explaining} to attack at each step. Instead, \cite{kos2017delving} proposed using the value function to determine when to launch attack. \cite{lin2017tactics} concentrated on attacking within specific steps of trajectories and employed a planner to craft perturbations that steer the agent toward a target state. \cite{behzadan2017vulnerability} explored the black-box setting, revealing transferable adversarial examples across different DQN models. In contrast to crafting perturbations solely based on the policy, \cite{pattanaik2018robust} introduced a more potent attack leveraging both the policy and the $Q$ function. Recently, \cite{zhang2020robust} formalized attacks on observations through a state-adversarial Markov decision process, demonstrating that the most powerful attacks can be learned as an RL problem. Based on this, \cite{zhang2021robust} and \cite{sun2021strongest} introduced the RL-based adversaries for black-box and white-box attacks, respectively.

\textbf{Robust Training for Deep RL.} To enhance the robustness of RL agents against adversarial attacks on observations, previous works have primarily focused on strategies involving adversarial examples during training. \cite{kos2017delving}; \cite{behzadan2017whatever} are concurrent works that first proposed to adversarially train DQN agents on Atari games. They used weak attacks on pixel space during rollouts and preserved perturbed frames for training. However, this approach exhibited limited improvements in several Atari games. Another line of research introduces a regularization-based approach to enhancing the robustness of DQN agents. \cite{russo2019optimal} proposed Lipschitz regularization, while \cite{zhang2020robust} used a hinge loss regularizer to promote the smoothness of $Q$ function under bounded perturbations. \cite{oikarinen2021robust} utilized robustness verification bound tools to compute the lower bound of $Q$ function, thereby certifying the robustness of action selection. In continuous control tasks, a similar adversarial training approach was initially explored by \cite{huang2017adversarial}, where attacks are induced from both policy and $Q$ function, and the trajectories sampled under attacks are used for training. However, recent work \cite{zhang2020robust} found that this approach may not reliably improve the robustness against new attacks. \cite{zhang2021robust} proposed an alternative training paradigm involving LSTM-based RL agents and a black-box RL-based adversary. Similarly, \cite{sun2021strongest} proposed the same training paradigm with the white-box RL adversary, leading to a more robust RL agent. Smoothness regularization has also been proposed to improve the robustness of the policy model in online RL setting \cite{shen2020deep,zhang2020robust}, as well as in offline setting \cite{yang2022rorl}.

\textbf{Input Transformation Based Defenses.} In the domain of image classification, aside from robust training methods \cite{madry2017towards,zhang2019theoretically}, there have been many studies on defending adversarial attacks through input transformations \cite{lu2017no,guo2018countering,xu2018feature,prakash2018deflecting,samangouei2018defense,gupta2019ciidefence,jin2019ape,salman2020denoised,nie2022diffusion}. Several works utilized traditional image processing such as image cropping \cite{guo2018countering}, rescaling \cite{lu2017no}, or bit depth reduction \cite{xu2018feature} to mitigate the impact of adversarial attacks on the classifier. Other methods employed the powerful generative models \cite{samangouei2018defense,jin2019ape,nie2022diffusion} or trained the denoisers \cite{liao2018defense,gupta2019ciidefence} to reconstruct clean images. However, given our focus on control tasks using RL, it is more appropriate to adopt cost-efficient techniques for countering attacks. Furthermore, denoisers composed of DNNs are also vulnerable to gradient-based attacks. Motivated by the utilization of image processing techniques, we propose to use VQ as an input transformation. Notably, unlike bit depth reduction \cite{xu2018feature} that employs uniform quantization, our method learns representative points to quantize inputs based on the statistic of input examples.

\textbf{Input Transformation on Deep RL.} Input transformation has been widely investigated in deep RL to enhance generalization \cite{tobin2017domain,cobbe2019quantifying} or improve sample efficiency \cite{laskin2020reinforcement,kostrikov2020image,luu2022utilizing}. Domain randomization, as proposed in \cite{tobin2017domain}, aims to transfer policies from simulators to the real world. Simple augmentations like cutout \cite{cobbe2019quantifying} or random convolution \cite{lee2020network}, as demonstrated in \cite{cobbe2019quantifying} and \cite{lee2020network}, have been shown to assist agents in generalizing to unseen environments. To reduce sample complexity in pixel-based RL, \cite{laskin2020reinforcement} applied image augmentations to the observations during agent training. Furthermore, \cite{kostrikov2020image} employed augmentation to regularize $Q$ functions, further improving sample efficiency. Vanilla vector quantization (VQ) has been utilized in several works to reduce the state space for generalization in tabular RL \cite{fernandez2000vqql, lau2002adaptive} and continuous control \cite{mavridis2021vector}. Our proposed method differs from these works by quantizing individual dimensions rather than entire vectors, making it more scalable. To the best of our knowledge, our use of input transformation represents the first attempt at leveraging it to enhance robustness against adversarial attacks in RL.

\section{PRELIMINARIES} \label{sec:preliminaries}

\subsection{Reinforcement Learning.} An reinforcement learning (RL) environment is modeled by a Markov decision process (MDP), defined as $\mathcal{M} = (\mathcal{S}, \mathcal{A}, R, P, \gamma)$, where $\mathcal{S}$ is the state space, $\mathcal{A}$ is the action space, $R: \mathcal{S}\times\mathcal{A}\times\mathcal{S} \rightarrow \mathbb{R}$ is the reward function, $P: \mathcal{S} \times \mathcal{A} \rightarrow \mathcal{S}$ is the transition probability distribution, and $\gamma \in [0, 1)$ is a discount factor. An agent takes actions based on a policy $\pi: \mathcal{S}\rightarrow \mathcal{A}$. The objective of the RL agent is to maximize the expected discounted return $\mathbb{E}_{\pi}[\sum_{t=0}^{\infty}\gamma^t R(s_t,a_t,s_{t+1})]$, which is the expected cumulative sum of rewards when following the policy $\pi$ in the MDP. 
This objective can be evaluated by a value function $V^{\pi}(s):=\mathbb{E}_{\pi}[\sum_{t=0}^{\infty}\gamma^tR_t|s_0=s]$, or the action value function $Q^{\pi}(s,a):=\mathbb{E}_{\pi}[\sum_{t=0}^{\infty}\gamma^tR_t|s_0=s, a_0=a]$. 

\subsection{Test-time Adversarial Attacks.} We consider adversarial attacks on the state observations during test time, which is formulated as SA-MDP \cite{zhang2020robust}. Specifically, during testing, the agent's observation is adversarially perturbed at every time step by an adversary equipped with a certain budget $\epsilon$. Note that, the adversary only alters the observations and the true underlying states of the environment do not change. This setting fits many realistic scenarios such as measurement errors, noise in sensory signals, or man-in-the-middle (MITM) attacks for a deep RL system. 
For example, in robotic manipulation, an attacker can add imperceptible noise to the camera capturing an object, however, the actual object's location is unchanged. 
In this paper, we consider a $\ell_{\infty}$ norm threat model, in which the adversary is restricted to perturb the observation $s$ into  $\hat{s} \in \mathcal{B}(s, \epsilon) = \{\hat{s}: \|s - \hat{s}\|_{\infty} \leq \epsilon\}$. Additionally, since the adversary only appears at the test time, we assume that the true states can be observed during training. This is important since our input transformation is learned to capture the statistic of states while training the agent.

\subsection{Vector Quantization.} Vector quantization (VQ) is a common technique widely used for learning discrete representation \cite{van2017neural,ozair2021vector,liu2021discrete,islam2022discrete}. In this work, we use VQ block similar to VQ-VAE \cite{van2017neural} with some modifications. We present the process of basic VQ here and leave the modification in the next section. Initially, the VQ block, referred to as  $\mathcal{Q}$, maintains a codebook $\mathbf{C}$ consisting of a set of items $ \{c_k\}_{k=1}^K $. Given an input vector $z$, $\mathcal{Q}$ outputs the item $c_m$ which is closest to $z$ in Euclidean distance as $c_m = \mathcal{Q}(z)$, with $m = \argmin_k \|c_k - z\|_{2}$.
The codebook item can be updated by $\ell_2$ error or the exponential moving average to move the item toward corresponding unquantized vectors assigned to that item. In the backward pass, the VQ block is treated as an identity function, referred
to as straight-through gradient estimation \cite{bengio2013estimating}. The hyper-parameter $K$ controls the size of the codebook, where lower values would lead to lossy compression but potentially yield more abstract representations.

\section{METHODOLOGY} \label{sec:method}

\subsection{Input Transformation based Defense for RL} \label{sec:vq_for_rl}

To understand the effectiveness of input transformation-based defense for the RL agent, we commence by analyzing its performance under adversarial perturbations utilizing the tools developed in SA-MDP \cite{zhang2020robust}. Let $f_1, f_2$ be functions mapping $\mathcal{S} \rightarrow \mathcal{S}$, and let $\pi$ represent a Gaussian policy with a constant independent variance. Assuming the policy network is $L$-Lipschitz continuous, we obtain:
\begin{equation}\label{eq:performance_bound}
\small
\max_{s\in\mathcal{S}}\{V^{\pi \circ f_1}(s)-V^{\pi\circ f_2 \circ \nu}(s)\} \leq \zeta \max_{s\in\mathcal{S}}\max_{\hat{s}\in \mathcal{B}(s, \epsilon)} \|f_1(s)-f_2(\hat{s})\|_2
\end{equation}
where, $\nu$ is optimal adversary corresponding to $\pi$, $\zeta$ is a constant independent of $\pi$. The proof of Eq. (\ref{eq:performance_bound}) relies on $L$-Lipschitz continuity of the policy network. 

Eq. (\ref{eq:performance_bound}) suggests two distinct approaches for narrowing the gap between natural performance and performance under perturbation. Firstly, when considering $f_1$ as an identity function, we can design $f_2$ to reconstruct $s$ from $\hat{s}$, which involves minimizing $\max_{\hat{s}\in \mathcal{B}(s, \epsilon)} \|s - f_2(\hat{s})\|_2$. Secondly, we can design $f_1$ and $f_2$ to reduce the difference in the transformed space, as expressed by $\max_{\hat{s}\in \mathcal{B}(s, \epsilon)} \|f_1(s) - f_2(\hat{s})\|_2$ being small. While the first approach can be achieved by utilizing a denoiser, it carries certain disadvantages, as discussed in Section \ref{sec:related_work}. Therefore, we opt for the second approach to counter adversarial attacks. In this approach, $V^{\pi\circ f_1}(s)$ is not guaranteed to be identical to $V^{\pi}(s)$, as the agent operates within the transformed space rather than the origin one. However, as long as $f_1$ retains the essential information from the original space, the agent's performance can be maintained, as shown in our experiments.

\subsection{VQ Mitigating Adversarial Perturbations} \label{sec:vq_counter_attack}

\begin{figure}[ht]
	\centering
	\begin{subfigure}{0.23\textwidth}
		\includegraphics[width=\textwidth]{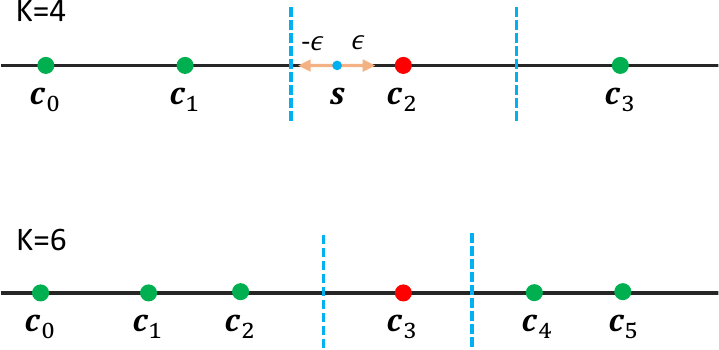}
		\caption{VQ reduces attack's space.}
		\label{fig:illustration_a}
	\end{subfigure}
	\begin{subfigure}{0.34\textwidth}
		\includegraphics[width=\textwidth]{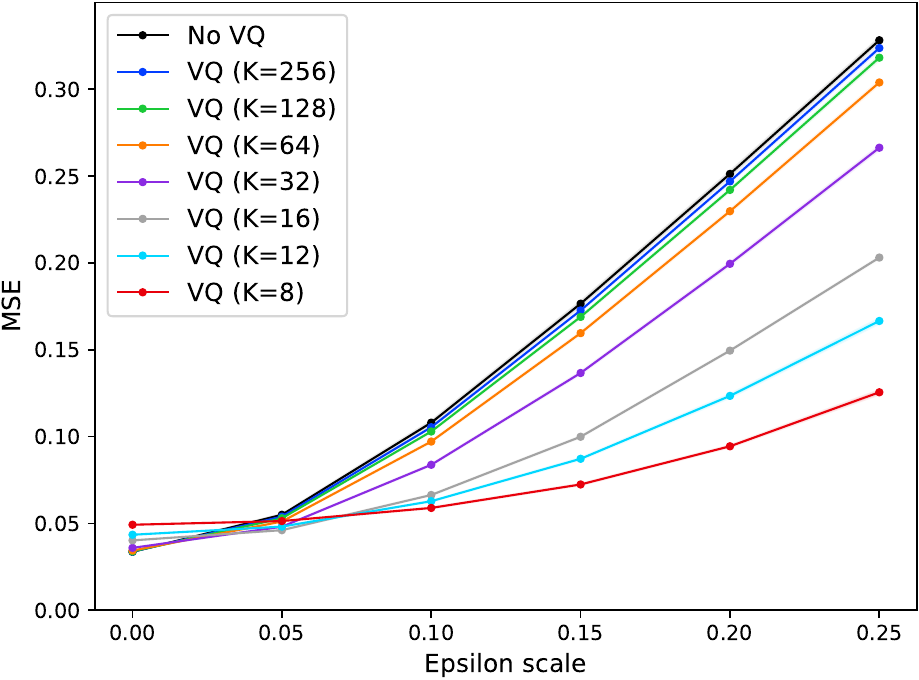}
		\caption{VQ as input transformation for regression.}
		\label{fig:illustration_b}
	\end{subfigure}
	\caption{(a) Illustration of using VQ to reduce space of adversarial attacks. The green and red dots indicate codebook items, whereas the red dot represents an item to which the state $s$ is assigned after VQ process. The blue dotted line indicates the boundaries. (b) Illustration of the effectiveness of VQ in countering attacks in the regression task.}
	\label{fig:illustration}
    \vspace{-0.5cm}
\end{figure}

Different to previous works \cite{van2017neural,liu2021discrete,islam2022discrete}, which use VQ to discretize the latent space, we directly apply VQ to each dimension of the raw input space as a transformation, and then train the RL agents directly on the transformed inputs. We define the space of adversarial attack in transformed space by $\bar{\mathcal{B}}(s, \epsilon) = \{\mathcal{Q}(\hat{s}): \|s - \hat{s}\|_{\infty} \leq \epsilon\}$. Intuitively, $\bar{\mathcal{B}}(s,\epsilon)$ is a set of possible items $c_k$ to which the perturbed state $\hat{s}$ can be assigned. We find that using VQ with an appropriate small codebook size as the input transformation effectively reduces the space of adversarial attacks, \ie, the size of $\bar{\mathcal{B}}$, without significantly reducing the natural performance of the agent. As depicted in Fig. \ref{fig:illustration_a} (top) for one-dimensional data, supposing the state $s$ is assigned to the item $c_2$, and the adversary can arbitrarily perturb the state $s$ within the $\epsilon$ ball. We can see that if the $\mathcal{B}(s, \epsilon)$ is still lying within the boundaries of $c_2$ (the blue dotted lines), $\mathcal{Q}$ will transform both $s$ and $\hat{s}\in \mathcal{B}(s, \epsilon)$ into the same item, \ie, $\mathcal{Q}(s) = \mathcal{Q}(\hat{s})$ for $\forall \hat{s} \in \mathcal{B}(s, \epsilon)$. Additionally, we also observe that the space of attacks is proportional to the size of the codebook. It means that $K$ decreases leading to smaller size of $\bar{\mathcal{B}}(s, \epsilon)$, thus stronger in resisting the adversarial perturbations. This is illustrated in Fig. \ref{fig:illustration_a}, larger $K$ shrinks the radius of items, while smaller $K$ enlarges the radius. Moreover, due to straight-through estimation, VQ also inherits non-differential properties as image transformations. For states lying close to the boundary, with appropriate small $K$ and not too large $\epsilon$, the transformed states are altered at most to the closest neighbor items.

To better understand the effectiveness of VQ for countering adversarial attacks, we illustrate it with a toy regression task. We train a predictor $\pi_{\theta}$ to regress from the state to the action on the \texttt{walker-medium-v2} dataset \cite{fu2020d4rl}, using VQ as input transformation. The model is optimized by minimizing the MSE between the prediction and ground truth action. At the test time, we introduce the adversary on the state $s$ to obtain the perturbed state $\hat{s} = \argmax_{\hat{s}\in \mathcal{B}(s,\epsilon)}\|\pi_{\theta}(s) - a\|_2$, with $a$ is the ground truth. The maximization is solved by using 10-step projected gradient descent (PGD) as in \cite{kurakin2016adversarial,pattanaik2018robust}. We evaluate the performance (\ie, MSE) under different scales of $\epsilon$, where $\epsilon=0$ corresponds to $\hat{s} = s$. 

As the result shown in Fig. \ref{fig:illustration_b}, smaller values of $K$ are more effective in countering the adversarial attacks, while the robustness of the model with larger $K$ will become closer to the model without using VQ. Additionally, we also observe that smaller $K$ will slightly decrease the natural performance (\ie, $\epsilon = 0$), which is resulted from the lossy compression. Therefore, we can adjust $K$ to control the trade-off between the natural performance and the robustness.

The VQ block originally uses the same codebook for all dimensions of the input \cite{van2017neural,ozair2021vector,liu2021discrete,islam2022discrete}. However, since we use VQ for the raw input observations instead of latent space, which is not learnable to adapt the codebook items, fitting the same codebook for every dimension might limit its expressiveness to approximate the state density, especially for the small size of the codebook. We propose to use separate codebooks for each dimension of input to improve its expressiveness. Therefore, our modified VQ block maintains a set of codebooks $\{\mathbf{C}^i\}_{i=1}^D$, where $D$ is the dimension of inputs, each codebook consists of a set of item $\{c_k\}_{k=1}^K$, with $c_k\in\mathbb{R}$. The codebook items are updated by $\ell_2$ error similar to \cite{van2017neural}. Specifically, considering the dimension $d$, let $\{s_{k,i}^{d}\}_{i=1}^{N_{d,k}}$ be the set of state elements closest to the item $c_k^d$, so that the codebook items are updated by minimizing following objective:
\begin{equation}\label{eq:vq_loss}
\small
\mathcal{L}_{VQ} = \frac{1}{D}\frac{1}{K}\sum_{d=1}^D\sum_{k=1}^K \frac{1}{N_{d,k}}\sum_{i=1}^{N_{d,k}}\|s_{k,i}^d - c_k^d\|_2^2
\end{equation}
While the VQ transformation described above is appealing for this approach, naively incorporating it into the RL algorithms may deteriorate the natural performance. This is because the state distribution induced by the policy is changed over the course of training, which is caused by the policy being continuously improved, thus the state distribution at the beginning may be very different at the end of training. As a result, the codebooks learned from the data induced by a low-performance policy at the beginning might be inadequate to reflect the state distribution induced by the high-performance policies at the end. Moreover, codebook items may converge to a local minimum similar to K-means \cite{bottou1994convergence}. To mitigate this, we propose to slowly update codebooks when the agent performance is low, \ie, when the agent is in exploration; and update faster when it reaches higher performance, \ie, when the agent is in exploitation. To control the rate of update, we scale $\mathcal{L}_{VQ}$ based on the current agent's performance. For simplicity, we approximate current performance by using the average value of $Q$ within a mini-batch during training. Then, we scale $\mathcal{L}_{VQ}$ in Eq. (\ref{eq:vq_loss}) by the factor $\lambda$ defined as follows:
\begin{equation}\label{eq:scale_factor}
\small
\lambda=\frac{\frac{1}{|B|}\sum_{s_i\in B}|Q^{\pi}(s_i, \pi(s_i))|}{\alpha} 
\end{equation}
where $\alpha$ is hyper-parameter, $\pi$ is the current training policy, $B$ is mini-batch. During training, we alternatively update between the RL agent and the codebooks.

\section{Experiments}

\subsection{Evaluation in MuJoCo}\label{sec:evaluation_online}

In this section, we evaluate the effectiveness of the proposed method against common adversarial attacks in both online and offline RL settings. We adopt three common attacks as used in \cite{zhang2020robust,zhang2021robust}, namely \textit{Random}, \textit{Action Diff}, and \textit{Min Q}. Given an attack budget $\epsilon$ and a state $s$, adversaries generate perturbed state $\hat{s}$ as follows: (1) \textit{Random}: $\hat{s}$ is uniformly sampled within $\mathcal{B}(s,\epsilon)$, (2) \textit{Action Diff}: $\hat{s}$ is induced from the agent's policy. Specifically, $\hat{s}$ is searched within $\mathcal{B}(s,\epsilon)$ to satisfy $\max_{\hat{s}\in\mathcal{B}(s,\epsilon)}D(\pi(\cdot|s)||\pi(\cdot|\hat{s}))$, with $D$ is KL divergence, (3) \textit{Min Q}: Different from \textit{Action Diff}, this adversary generates perturbations based on both the agent's policy and $Q$ function, which is a relatively stronger attack. $\hat{s}$ is selected to satisfy $\min_{\hat{s}\in\mathcal{B}(s,\epsilon)}Q(s, a)$, with $a$ is the policy output. For \textit{Action Diff} and \textit{Min Q}, we use 10-step PGD as in \cite{kurakin2016adversarial,pattanaik2018robust}. 

\subsubsection{Online RL} For online RL setting, we follow the experiment setup as in \cite{zhang2020robust}, but use Soft Actor-Critic (SAC) as a base RL algorithm due to its sample efficiency and higher performance on Gym MuJoCo \cite{haarnoja2018soft}. We conduct experiments on five environments including Walker2d, Hopper, Ant, Reacher, and InvertedPendulum. For $\alpha$ in Eq. (\ref{eq:scale_factor}), we search in $\{30, 40, 50, 60\}$ and set a value of $30$ for Hopper and $60$ for the others. Regarding the codebook size, we set $K = 8$ for Walker2d and Reacher, and $K=16$ for the others. We compare the proposed method with vanilla SAC and SAC with bit depth reduction \cite{xu2018feature} (SAC-BDR), where uniform quantization is naively performed in each dimension of the input. Additionally, we also incorporate VQ into a strong adversarial training baseline \cite{zhang2020robust}, referred to as SAC-SA, to see whether it can further improve the robustness.

We evaluate the robustness of methods under different scales of $\epsilon$, where $\epsilon=0$ corresponds to the natural performance (see in Appendix), and generate the return curves under different attack levels. For better quantitative measurement, we consider the \textit{robustness score} (RC) defined as the areas under the perturbation curve, where the returns are normalized between $R_{min}$ and $R_{max}$. Specifically, given a normalized perturbation curve, the RC is computed by $RC = \frac{1}{N}\sum_{i=1}^N \mathbf{R}[i]$, with $\mathbf{R}$ is the list of returns evaluated on $N$ monotonically increasing attack scales, each return value is averaged over 50 episodes similar to \cite{zhang2020robust}. 

\begin{table}[t]
\renewcommand{\arraystretch}{0.8}
\centering
\caption{Average robustness score and standard deviation in Mujoco for online RL setting over five seeds. In each environment, we bold the highest average score in each setting: with and without robust training-based defense.}\label{tab:aupc_online}
\begin{tabular}{l@{\hspace{3pt}}l@{\hspace{4pt}}c@{\hspace{5pt}}c@{\hspace{5pt}}c@{\hspace{5pt}}c@{\hspace{5pt}}}
    \toprule
    Env.   & Method  &   Random &  Action Diff  &  Min Q   &  Average \\
    \midrule
    \multirow{6}{*}{\rotatebox[origin=c]{90}{\text{Walker2d}}}
    & SAC       &     $84.4 \pm 5$ &     $42.1 \pm 4$ &   $40.4 \pm 3$ &      $55.7$  \\
    & SAC-BDR   &     $88.8 \pm 4$ &     $48.6 \pm 6$ &   $43.8 \pm 7$ &      $60.4$ \\		
    & SAC-VQ    &     $88.9 \pm 4$ &     $65.4 \pm 8$ &   $50.9 \pm 7$ &      $\textbf{68.4}$  \\
    \cmidrule(r){2-6}
    & SAC-SA    &     $91.6 \pm 1$ &     $66.3 \pm 7$ &   $47.7 \pm 7$ &      $68.6$  \\
    & SAC-SA-BDR&     $91.1 \pm 2$ &     $72.1 \pm 2$ &   $50.9 \pm 3$ &      $71.4$  \\
    & SAC-SA-VQ &     $92.5 \pm 5$ &     $77.5 \pm 6$ &   $52.9 \pm 6$ &      $\textbf{74.3}$  \\
    \midrule
    
    \multirow{6}{*}{\rotatebox[origin=c]{90}{\text{Hopper}}}   
    & SAC       &     $65.2 \pm 7$ &     $36.8 \pm 5$ &   $36.4 \pm 4$ &      $46.1$  \\
    & SAC-BDR   &     $72.7 \pm 5$ &     $45.7 \pm 6$ &   $44.0\pm 3$ &      $54.1$  \\
    & SAC-VQ    &     $72.5 \pm 6$ &     $50.2 \pm 5$ &   $45.5 \pm 5$ &      $\textbf{56.2}$  \\
    \cmidrule(r){2-6}
    & SAC-SA    &     $91.7 \pm 3$ &     $67.1 \pm 4$ &   $61.8 \pm 4$ &      $73.5$  \\
    & SAC-SA-BDR&     $90.3 \pm 3$ &     $68.6 \pm 4$ &   $ 63.55\pm 4$ &      $74.2$  \\
    & SAC-SA-VQ &     $91.2 \pm 8$ &     $70.2 \pm 7$ &   $64.42 \pm 5$ &      $\textbf{75.3}$  \\
    \midrule
    
    \multirow{6}{*}{\rotatebox[origin=c]{90}{\text{Ant}}}   
    & SAC       &     $71.5 \pm 4$ &     $30.8 \pm 3$ &   $32.02 \pm 3$ &      $44.8$  \\
    & SAC-BDR   &     $68.0 \pm 6$ &     $34.9 \pm 4$ &   $31.12 \pm 5$ &      $44.7$  \\
    & SAC-VQ    &     $78.5 \pm 8$ &     $41.9 \pm 7$ &   $39.34 \pm 6$ &      $\textbf{53.2}$  \\
    \cmidrule(r){2-6}
    & SAC-SA    &     $85.3 \pm 3$ &     $50.3 \pm 7$ &   $53.73 \pm 5$ &      $63.1$  \\
    & SAC-SA-BDR&     $81.5 \pm 4$ &     $54.6 \pm 5$ &   $52.06 \pm 6$ &      $62.7$  \\
    & SAC-SA-VQ &     $86.7 \pm 3$ &     $59.7 \pm 4$ &   $55.15 \pm 2$ &      $\textbf{67.2}$  \\
    \midrule
    
     \multirow{6}{*}{\rotatebox[origin=c]{90}{\text{Reacher}}}   
     & SAC      &     $99.2 \pm 0.2$ &     $94.9 \pm 0.7$ &   $95.27 \pm 0.6$ &      $96.5$  \\
     & SAC-BDR  &     $99.2 \pm 0.2$ &     $96.4 \pm 0.6$ &   $96.36 \pm 0.4$ &      $97.3$ \\
     & SAC-VQ   &     $99.3 \pm 0.1$ &     $97.2 \pm 0.1$ &   $96.89 \pm 0.2$ &      $\textbf{97.8}$  \\
    \cmidrule(r){2-6}
     & SAC-SA    &     $99.6 \pm 0.1$ &     $98.1 \pm 0.2$ &   $97.02 \pm 0.3$ &     $98.3$  \\
     & SAC-SA-BDR&     $99.2 \pm 0.4$ &     $98.1 \pm 0.5$ &   $97.81 \pm 0.5$ &     $98.4$   \\
     & SAC-SA-VQ &     $99.6 \pm 0.1$ &     $98.9 \pm 0.1$ &   $97.92 \pm 0.2$ &     $\textbf{98.8}$  \\
     \midrule
        
     \multirow{6}{*}{\rotatebox[origin=c]{90}{\text{Pendulum}}}   
     & SAC       &     $88.2 \pm 12$ &     $60.2 \pm 11$ &   $69.28 \pm 13.4$ &     $72.5$  \\
     & SAC-BDR   &     $99.8 \pm 3$  &     $80.5 \pm 10$  &   $88.47 \pm 6.9$  &     $89.6$  \\
     & SAC-VQ    &     $100 \pm 0$   &     $91.1 \pm 4$  &   $94.98 \pm 2.1$  &     $\textbf{95.3}$  \\
    \cmidrule(r){2-6}
     & SAC-SA    &     $100 \pm 0$ &     $88.3 \pm 1$ &   $62.49 \pm 3.7$ &      $83.6$  \\
     & SAC-SA-BDR&     $100 \pm 0$ &     $92.2 \pm 3$ &   $58.91 \pm 4.6$ &      $83.7$  \\
     & SAC-SA-VQ &     $100 \pm 0$ &     $97.8 \pm 1$ &   $59.75 \pm 7.4$ &      $\textbf{85.8}$  \\
    \bottomrule
\end{tabular}
\vskip -0.3in
\end{table}

According to the result shown in Tab. \ref{tab:aupc_online}, our method consistently enhances the robustness score over vanilla SAC, with an averaged improvement of 11\%. While SAC-BDR succeeds in enhancing the robustness of SAC, it falls short by 5\% compared to ours, highlighting the advantage of learning codebooks over uniform quantization. When coupled with SAC-SA, the robustness can be further enhanced. However, in the case of the InvertedPendulum under the \textit{Min Q} attack, VQ exacerbates performance degradation. This could be attributed to the use of smoothness regularization, which also negatively impacts performance compared to vanilla SAC. Therefore, the combination with VQ may degrade performance. Remarkably, SAC-VQ achieves comparable performance with SAC-SA in the Walker2d environment and surpasses it in the InvertedPendulum. Overall, utilizing VQ as input transformation effectively mitigates the impact of attacks and further enhances the robustness of robust training-based defenses.

\subsubsection{Offline RL} For offline RL setting, we use TD3BC \cite{fujimoto2021minimalist} as a baseline and conduct on \{walker2d, hopper, ant\}-medium-v2 datasets. Following a similar comparison in the online setting, we incorporate VQ into a robust training-based defense \cite{zhang2020robust,yang2022rorl}. We search for $K$ within $\{8, 12, 16\}$ for all datasets, except for hopper-medium-v2 where we broaden the search to $\{8, 16, 24, 28, 32\}$ and report the best score. This setting presents more challenges compared to the online setting due to the state distribution shift problem \cite{levine2020offline}. This problem arises because codebooks learned from the offline dataset might inaccurately reflect the density of states induced by the current policy at test time. We investigate whether VQ is helpful for improving the robustness in this challenging setting. The evaluation metric is same with online setting. As shown in Tab. \ref{tab:aupc_offline}, we observe that VQ is still able to enhance the robustness of base algorithms in almost tasks. However, the magnitude of improvements is lower compared to the online setting. The most substantial improvement are observed in the Walker2d and Ant datasets. 

\begin{table}[t]
\renewcommand{\arraystretch}{0.8}
\centering
\caption{Average robustness score and standard deviation in Mujoco for offline RL setting over five seeds. In each environment, we bold the highest average score in each setting: with and without robust training-based defense}\label{tab:aupc_offline}
\begin{tabular}{l@{\hspace{4pt}}l@{\hspace{5pt}}c@{\hspace{5pt}}c@{\hspace{5pt}}c@{\hspace{5pt}}c@{\hspace{5pt}}}
    \toprule
    Env.   &  Method &   Random &  Action Diff  &  Min Q   &  Average \\
    \midrule
    \multirow{4}{*}{\rotatebox[origin=c]{90}{\text{Walker2d}}}
    & TD3BC        &     $81.6 \pm 2.2$ &     $57.7 \pm 3.2$ &   $38.4 \pm 2$ &      $59.3$  \\
    &  TD3BC-VQ    &     $83.0 \pm 1.3$ &     $70.7 \pm 1.6$ &   $47.8 \pm 4$ &      $\textbf{67.2}$  \\
    \cmidrule(r){2-6}
    &  TD3BC-SA    &     $84.4 \pm 1.3$ &     $68.9 \pm 2$ &   $43.0 \pm 3.4$ &      $65.5$  \\
    &  TD3BC-SA-VQ &     $85.0 \pm 0.4$ &     $74.3 \pm 1.8$ &   $45.1 \pm 3.3$ &      $\textbf{68.1}$  \\
    \midrule
    
    \multirow{4}{*}{\rotatebox[origin=c]{90}{\text{Hopper}}}   
    &  TD3BC       &     $53.8 \pm 0.8$ &     $43.7 \pm 1.1$ &   $21.0 \pm 3.1$ &      $39.6$  \\
    &  TD3BC-VQ    &     $52.2 \pm 1.2$ &     $44.2 \pm 0.7$ &   $27.6 \pm 2.9$ &      $\textbf{41.4}$  \\
    \cmidrule(r){2-6}
    &  TD3BC-SA    &     $54.4 \pm 0.9$ &     $45.7 \pm 0.6$ &   $23.5 \pm 1.5$ &      $\textbf{41.2}$  \\
    &  TD3BC-SA-VQ &     $52.6 \pm 4.9$ &     $44.5 \pm 3$   &   $26.2 \pm 1.6$ &      $41.1$  \\	
    \midrule
    \multirow{4}{*}{\rotatebox[origin=c]{90}{\text{Ant}}}   
    &  TD3BC       &     $84.8 \pm 5.3$ &     $52.6 \pm 5.6$ &   $56.3 \pm 4.5$ &      $64.6$ \\
    &  TD3BC-VQ    &     $83.0 \pm 7.2$ &     $58.8 \pm 4.9$ &   $60.5 \pm 4.7$ &      $\textbf{67.5}$ \\
    \cmidrule(r){2-6}
    &  TD3BC-SA    &     $88.7 \pm 4.1$ &     $60.2 \pm 3.6$ &   $61.9 \pm 4.1$ &      $70.3$ \\
    &  TD3BC-SA-VQ &     $92.7 \pm 3.2$ &     $70.7 \pm 3.1$ &   $68.8 \pm 1.4$ &      $\textbf{77.4}$ \\
    \bottomrule
\end{tabular}
\vskip -0.25in
\end{table}

\begin{figure*}[ht]
\centering
\includegraphics[width=0.49\textwidth]{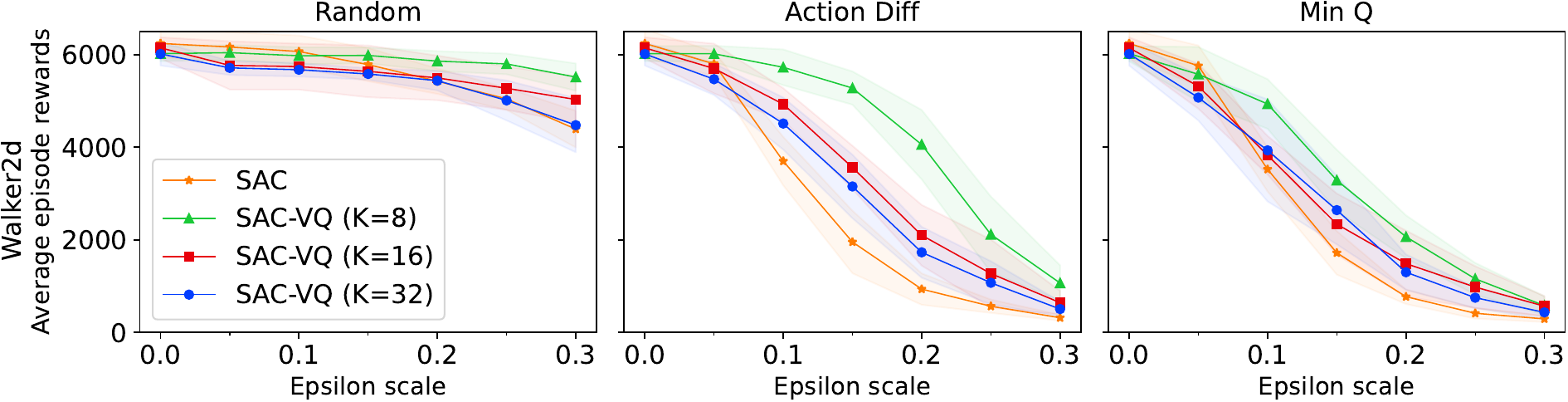}
\includegraphics[width=0.49\textwidth]{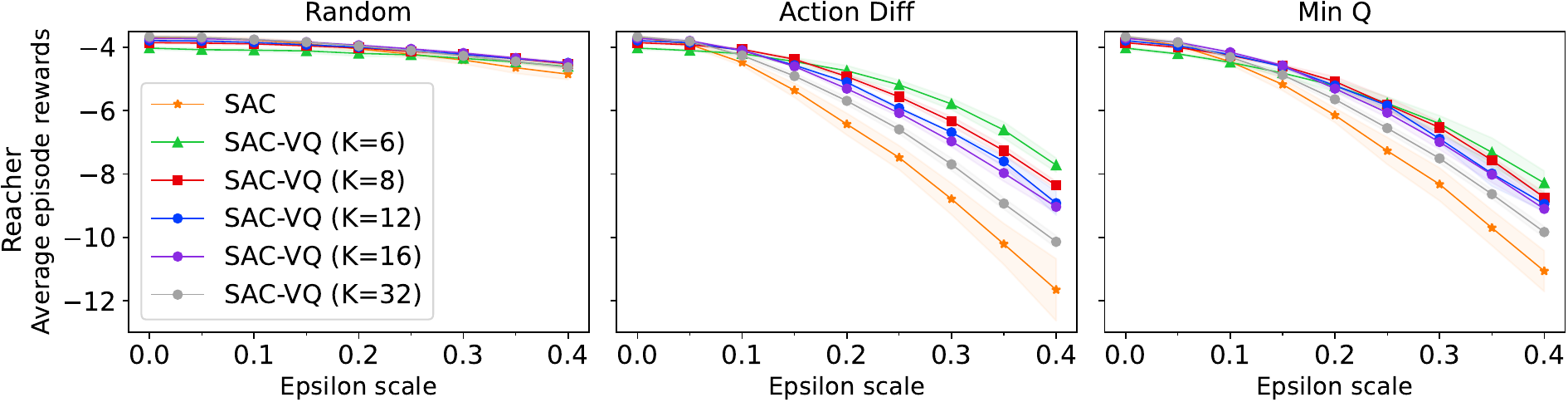}
\caption{The comparison between agents using different sizes of the codebook on Walker2d and Reacher.}
\label{fig:varying_k}
\vskip -0.2in
\end{figure*}

\begin{figure*}[]
\centering
\includegraphics[width=0.49\textwidth]{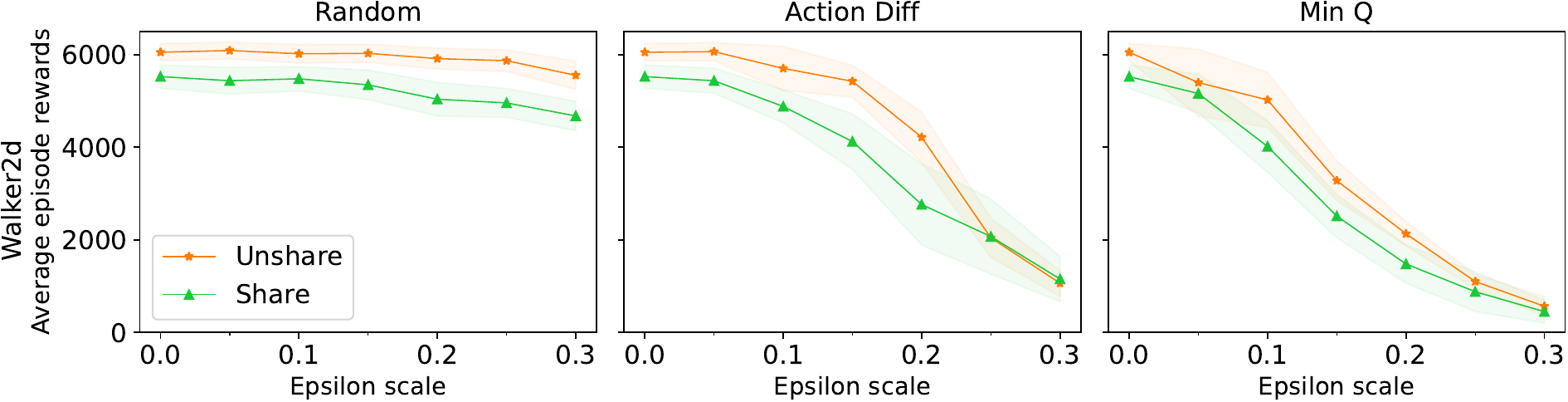}
\includegraphics[width=0.49\textwidth]{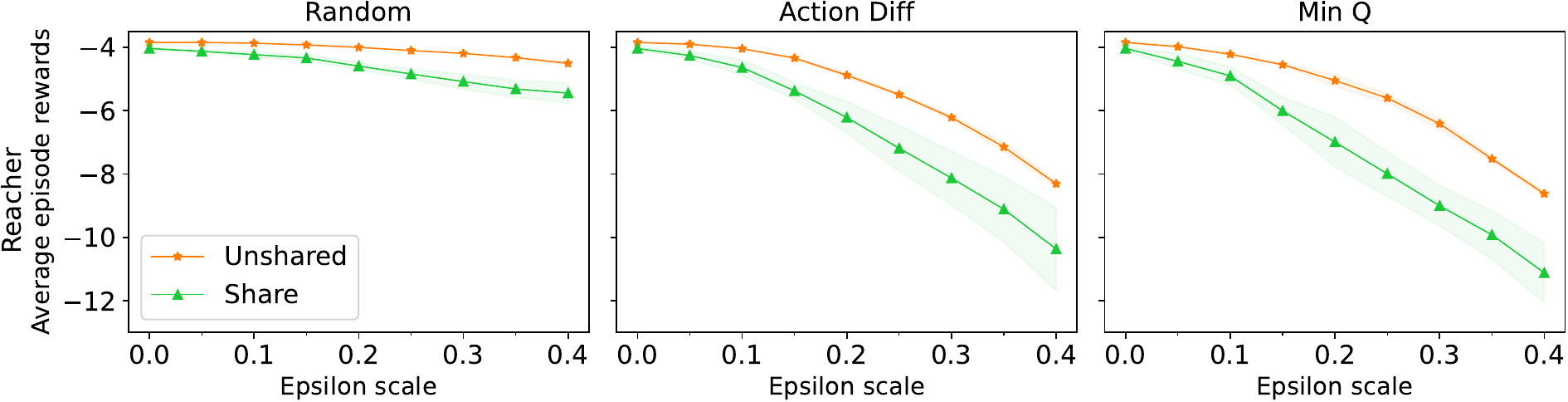}
\caption{The comparison between sharing and separate codebooks for all dimensions of states on Walker2d and Reacher.}
\label{fig:share_unshare_codebook}
\vskip -0.15in
\end{figure*}

\subsection{Evaluation in Atari}

We investigate effectiveness of VQ into the Double DQN \cite{van2016deep} on two Atari games: Freeway and Pong. These environments feature high-dimensional pixel inputs and discrete action spaces. We set $K=4$ for these environments. For the robustness evaluation, we use 10-step $l_{\infty}$-PGD untargeted attack. Additionally, we  also incorporate  bit depth reduction for the DQN agent, referred as DQN-BDR. Furthermore, we integrate the proposed method into a state-of-the-art method, namely RADIAL \cite{oikarinen2021robust}, to investigate whether it can enhance robustness. As demonstrated in Tab. \ref{tab:atari_vq}, the DQN-VQ is more effective compared to DQN-BDR. This outcome underscores the advantages of learning codebooks over uniform quantization. Surprisingly, our method is able to achieve comparable with RADIAL without adversarial training. By combining VQ with RADIAL, we achieve further improvement in robustness, especially at larger values of $\epsilon$ such as 10/255 for Pong and 5/255 for Freeway.

\begin{table}[t]
\renewcommand{\arraystretch}{0.9}
\centering
\caption{The mean reward of 10 runs $\pm$ the standard error in the Atari domain. Each run is averaged over 20 episodes. The highest score in each column of environments is bold.}\label{tab:atari_vq}
    \begin{tabular}{l@{\hspace{5pt}}l@{\hspace{5pt}}c@{\hspace{5pt}}c@{\hspace{5pt}}c@{\hspace{5pt}}}
    \toprule
    Env.   & Method/Metric     &   Natural Reward &  \multicolumn{2}{c}{PGD} \\
    \midrule
    \multicolumn{2}{l}{$\epsilon$}    & 0  &    3/255 & 5/255  \\
    \midrule
    \multirow{6}{*}{\rotatebox[origin=c]{90}{\text{Freeway}}} 
    & DQN       &  \textbf{33.9 $\pm$ 0.07}   &  0.0 $\pm$ 0.0  & 0.0 $\pm$ 0.0      \\
    & DQN-BDR   &  33.2 $\pm$ 0.1  & 32.7 $\pm$ 0.6  & 27.8 $\pm$ 0.7      \\
    & DQN-VQ    &  33.5 $\pm$ 0.7  & 32.9 $\pm$ 0.8  & 28.1 $\pm$ 1.7      \\
    \cmidrule(r){2-5}
    & RADIAL    &  33.4 $\pm$ 0.7  & \textbf{33.4 $\pm$ 0.6}   &  28.5 $\pm$ 0.9    \\
    & RADIAL-BDR&  \textbf{33.9 $\pm$ 0.9}  & 33.0 $\pm$ 0.8            &  30.5 $\pm$ 0.9    \\
    & RADIAL-VQ &  \textbf{33.9 $\pm$ 0.4}  & 33.2 $\pm$ 0.7   &  \textbf{32.4 $\pm$ 1.1}      \\
    \midrule
    \multicolumn{2}{l}{$\epsilon$}    & 0  &   5/255 & 10/255 \\
    \midrule
    \multirow{6}{*}{\rotatebox[origin=c]{90}{\text{Pong}}} 
    & DQN       & -21.0 $\pm$ 0.0 &   -21.0 $\pm$ 0.0  & -21.0 $\pm$ 0.0   \\
    & DQN-BDR   & 20.9 $\pm$ 0.4  &  14.7 $\pm$ 5.5  & -21.0 $\pm$ 0.0  \\
    & DQN-VQ    & \textbf{21.0 $\pm$ 0.0}  &  20.4 $\pm$ 0.9   & -20.3 $\pm$ 0.5 \\
    \cmidrule(r){2-5}
    & RADIAL    & \textbf{21.0 $\pm$ 0.0}  &  \textbf{21.0 $\pm$ 0.0}   & -20.8 $\pm$ 0.4  \\
    & RADIAL-BDR&  \textbf{21.0 $\pm$ 0.0}       & \textbf{21.0 $\pm$ 0.0}            &  16.5 $\pm$ 0.9    \\
    & RADIAL-VQ & \textbf{21.0 $\pm$ 0.0}  &  \textbf{21.0 $\pm$ 0.0} & \textbf{21.0 $\pm$ 0.0}    \\
    \midrule
    \end{tabular}
    \vskip -0.3in
\end{table}

\subsection{Ablation Study}\label{sec:ablation}

\textit{Effectiveness of Codebook Size.} We provide the experiment showing the effectiveness of different codebook sizes in Fig. \ref{fig:varying_k}. Across environments, the small values of $K$ often lead to more robustness in a wide range of attack scales. However, too small $K$ causes a drop in natural performance, and the large value of $K$ (e.g., 32) tends to have little benefit for resisting perturbations as analyzed in Section \ref{sec:vq_counter_attack}. It is important to emphasize that the dynamics of different environments can vary significantly, which in turn affects the distribution of states. Therefore, selecting an appropriate value of $K$ for each environment is crucial to achieving robust performance.

\textit{Effectiveness of Separate Codebooks.} The Fig. \ref{fig:share_unshare_codebook} compares performance between shared and unshared (\ie, separate) codebook for each dimension. The result shows that using the shared codebook across dimensions can reduce natural performance due to its limited expressiveness. Consequently, the robustness under attacks is also decreased. 

\begin{figure}[ht]
    \centering
    \vskip -0.05in
    \includegraphics[width=0.2\textwidth]{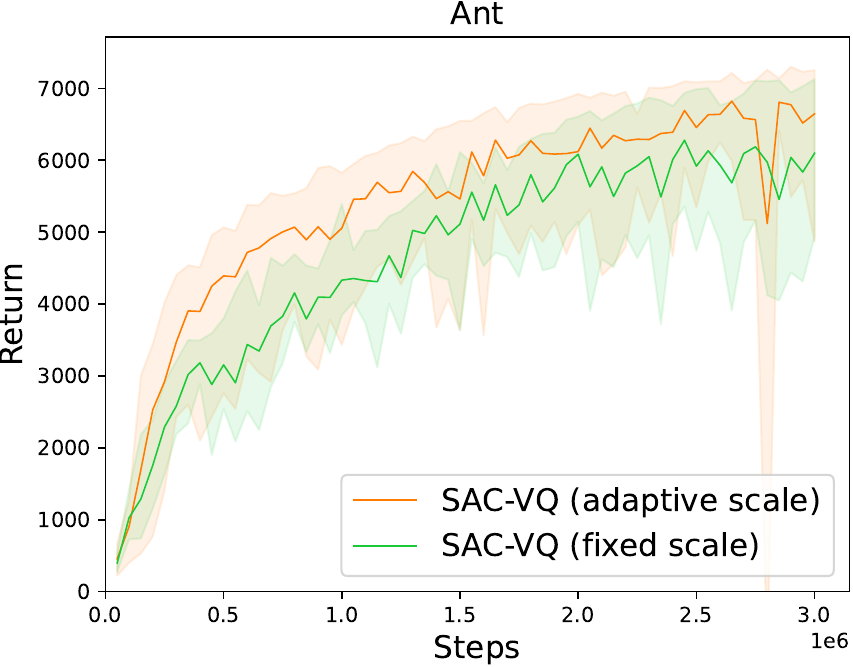}
    \includegraphics[width=0.2\textwidth]{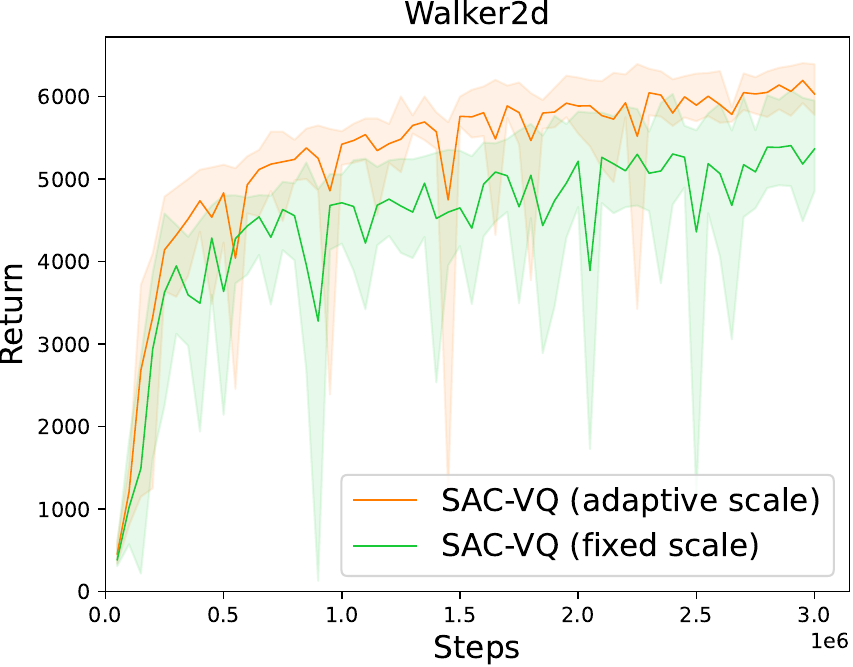}
    \caption{Ablation on adaptive learning codebook.}
    \label{fig:adaptive_scale}
    \vskip -0.15in
\end{figure}

\textit{Adaptive Learning Codebook.} To demonstrate the effectiveness of adaptive scale during updating codebooks, we show the natural performance during training of Walker2d and Ant in Fig. \ref{fig:adaptive_scale}. The ``\textit{fixed scale}'' means no scale used for $\mathcal{L}_{VQ}$ loss. The result shows that adaptive scale is important to achieve high natural performance at the end of training.

\begin{figure}[ht]
    \centering
    \vskip -0.05in
    \includegraphics[width=0.2\textwidth]{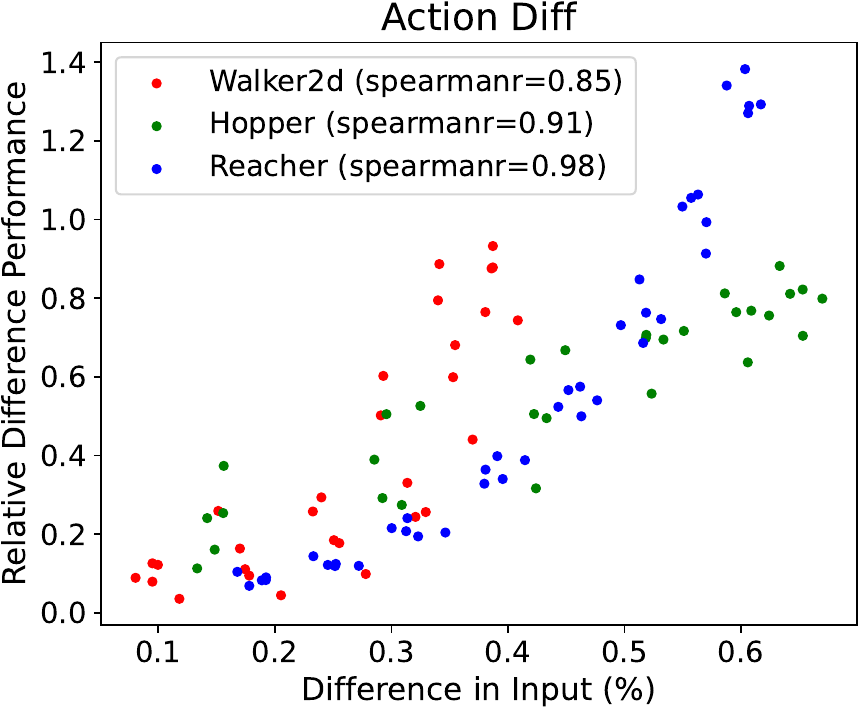}
    \includegraphics[width=0.2\textwidth]{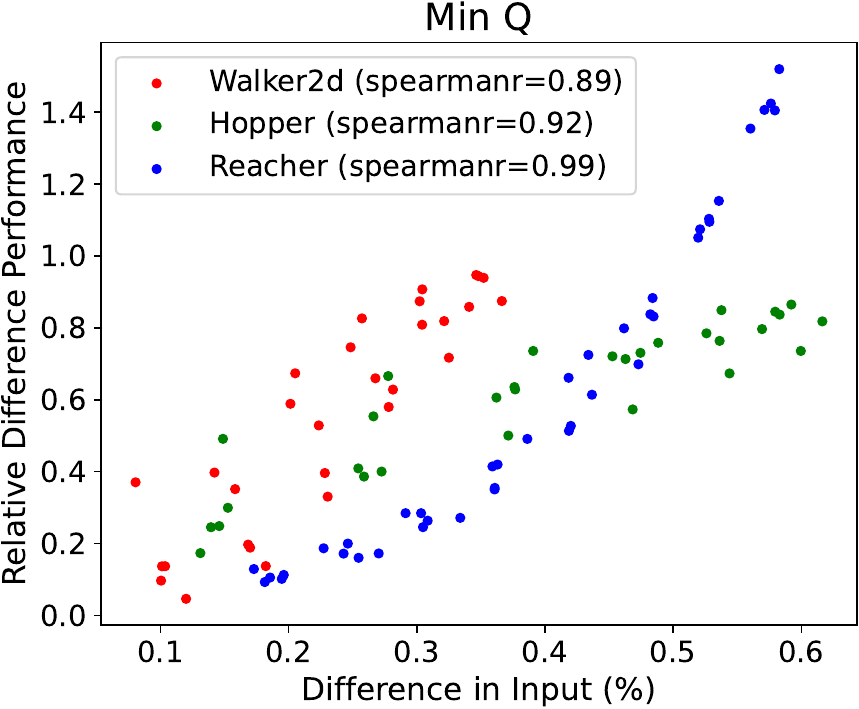}
    \caption{The correlation between the input difference and relative difference of performance.}
    \label{fig:spearmanr}
    \vskip -0.15in
\end{figure}

\textit{Input Difference vs. Robustness.} We measure Spearman's rank correlation coefficient between (1) the difference between clean and perturbed inputs after quantized and (2) the relative difference between the natural and robust performance under perturbation. The result shown in Fig. \ref{fig:spearmanr} indicates a high correlation between the two quantities. This supports Eq. (\ref{eq:performance_bound}) that decreasing the input difference will increase its robustness performance.

\begin{table}[ht]
\renewcommand{\arraystretch}{0.9}
\centering
\vskip -0.1in
\caption{Average training time on Walker2d-v2.}
\begin{tabular}{@{}lc@{}}
    \toprule
    Method & Runtime (s/iteration)  \\ 
    \midrule
    SAC         & 0.0220     \\
    SAC-VQ      & 0.0232     \\
    SAC-SA      & 0.0326     \\
    SAC-SA-VQ   & 0.0350     \\
    \bottomrule
\end{tabular}
\label{tab:computation_cost}
\vskip -0.2in
\end{table}

\subsection{Computational Cost Comparison.} We compare training time when using VQ transformation on a single machine with one GPU (RTX 3080). The result is shown in Tab. \ref{tab:computation_cost}. When combined with vanilla SAC and SAC-SA, VQ slightly increases the training time to $5\%$ and $7\%$, respectively.

\section{Conclusion} We have presented a novel defense based on input transformation to counter adversarial attacks on state observations. Our proposed approach is both cost-efficient and highly effective in defending against such attacks. Furthermore, when combined with robust training-based defenses, it significantly enhances the overall robustness of RL agents. We have conducted thorough analyses to evaluate the efficacy of vector quantization in countering attacks. To the best of our knowledge, this is the first study to investigate the use of input transformation-based defenses for RL. 

\bibliographystyle{plain} 
\bibliography{IEEEfull,references}

\section*{APPENDIX}

\noindent
\textbf{Performance bound.} SA-MDP \cite{zhang2020robust} provides an upper bound of performance gap between the two policies trained on non-adversarial MDP and state-adversarial MDP (SA-MDP), respectively. We based on this to derive our upper bound in Eq. (\ref{eq:performance_bound}). Formally, given a policy $\pi$ and its value function $V^{\pi}(s)$, under the optimal adversary $\nu$ in SA-MDP, theorem 5 in \cite{zhang2020robust} stated that:
\begin{equation}\label{eq:samdp_bound}
\small
\max_{s\in\mathcal{S}}\{V^{\pi}(s)-V^{\pi\circ\nu}(s)\} \leq \kappa \max_{s\in\mathcal{S}}\max_{\hat{s}\in \mathcal{B}(s,\epsilon)}D_{TV}(\pi(\cdot|s), \pi(\cdot|\hat{s}))
\end{equation}
where, $D_{TV}(\pi(\cdot|s), \pi(\cdot|\hat{s}))$ is the total variance distance between $\pi(\cdot|s)$ and $\pi(\cdot|\hat{s})$, $\kappa$ is a constant that does not depend on $\pi$, $\mathcal{B}(s,\epsilon)=\{\hat{s}: \|s-\hat{s}\|_{\infty}\leq \epsilon\}$, and $\pi\circ\nu$ denotes the policy under perturbations: $\pi(a|\nu(s))$. The total variation distance is not easy to compute for most distributions, thus we upper bound $D_{TV}$ by the \textit{KL} divergence:
\begin{equation}\label{eq:tv_kl}
    \small
    D_{TV}(\pi(\cdot|s), \pi(\cdot|\hat{s})) \leq \sqrt{\frac{1}{2}KL\left(\pi(\cdot|s) \parallel \pi(\cdot|\hat{s})\right)}.
\end{equation}
We assume that the policy is Gaussian with constant independence variance, which is commonly used in RL algorithms such as TD3 \cite{fujimoto2018addressing}. Supposing that $\pi(\cdot|s) \sim \mathcal{N}(\mu_s,\Sigma_s)$ and $\pi(\cdot|\hat{s}) \sim \mathcal{N}(\mu_{\hat{s}},\Sigma_{\hat{s}})$, where $\mu \in \mathbb{R}^d$ and $\mu_s, \mu_{\hat{s}}$ are respectively produced by neural networks $\mu_{\theta}(s), \mu_{\theta}(\hat{s}) $, and $\Sigma$ is a diagonal matrix independent of state $s$ (\ie, $\Sigma_s = \Sigma_{\hat{s}} = \Sigma$). Assuming policy network is $L$-Lipschitz continuous, we have:

\begin{equation}\label{eq:kl_lipschitz}
    \small
    \begin{aligned}
        KL(\pi(\cdot|s) \parallel \pi(\cdot|\hat{s})) &=\frac{1}{2}(\log\frac{|\Sigma_{\hat{s}}|}{|\Sigma_s|} - d + tr(\Sigma_{\hat{s}}^{-1}\Sigma_s) \\
        & \hspace{2cm} + (\mu_{\hat{s}} - \mu_{s})^{\top}\Sigma_{\hat{s}}^{-1}(\mu_{\hat{s}} - \mu_{s})) \\
        &\leq C^2\|\mu_{\hat{s}} - \mu_{s}\|^2_2 \\
        &= C^2\|\mu_{\theta}(s) - \mu_{\theta}(\hat{s})\|_2^2\\
        &\leq C^2L\|s - \hat{s}\|_2^2 .
    \end{aligned}
\end{equation}
The first inequality because of $\Sigma_{\hat{s}}, \Sigma_{s}$ are positive, thus exist $C\in\mathbb{R}^+$ to satisfy this. The second inequality because of the policy is $L$-Lipschitz continuous. Now we introduce two function $f_1, f_2$ that map $\mathcal{S}\rightarrow\mathcal{S}$, and apply these two functions into input states $s$ and $\hat{s}$, respectively. We have:
\begin{equation}\label{eq:transform}
\small
KL\left(\pi(\cdot|f_1(s)) \parallel \pi(\cdot|f_2(\hat{s}))\right) \leq C^2L\|f_1(s) - f_2(\hat{s})\|_2^2.
\end{equation}
Combining (\ref{eq:samdp_bound}), (\ref{eq:tv_kl}), (\ref{eq:kl_lipschitz}), and (\ref{eq:transform}) we have:
\begin{equation}
\small
\begin{aligned}
    \max_{s\in\mathcal{S}}\{V^{\pi\circ f_1}(s) -V^{\pi\circ f_2\circ\nu}(s)\} \leq \zeta\max_{s\in\mathcal{S}}\max_{\hat{s}\in \mathcal{B}(s,\epsilon)}\|f_1(s)-f_2(\hat{s})\|_2
\end{aligned}
\end{equation}
where, $\zeta=\frac{1}{\sqrt{2}}\kappa C\sqrt{L}$,  $\pi\circ f_1$ and $\pi\circ f_2\circ\nu$ denote $\pi(\cdot|f_1(s))$ and $\pi(\cdot|f_2(\nu(\hat{s})))$, respectively. $\hfill \square$

\begin{figure}[ht]
    \centering
    \includegraphics[width=0.5\textwidth]{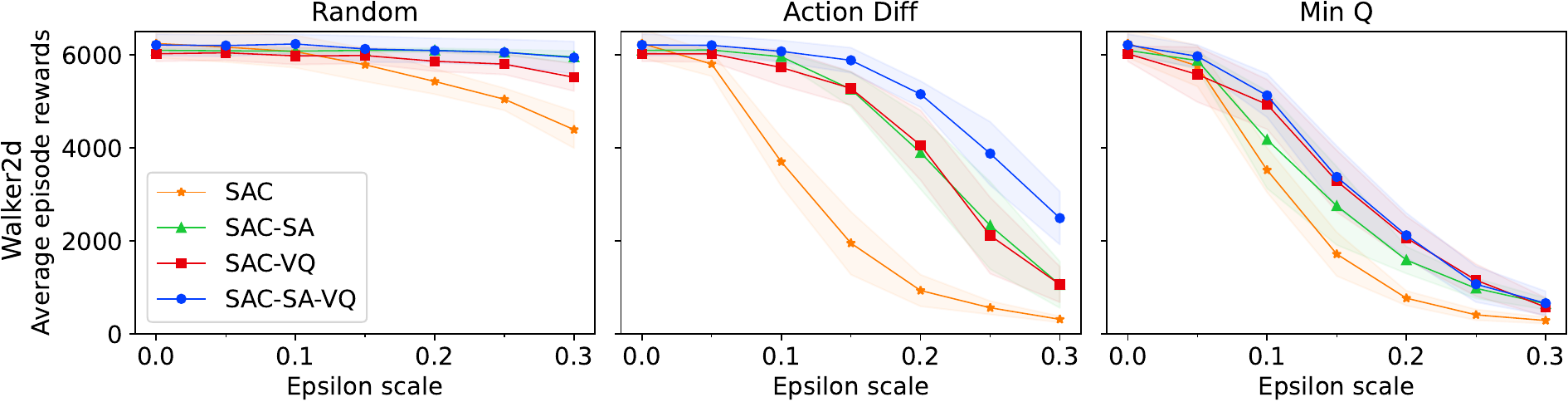}
    \includegraphics[width=0.5\textwidth]{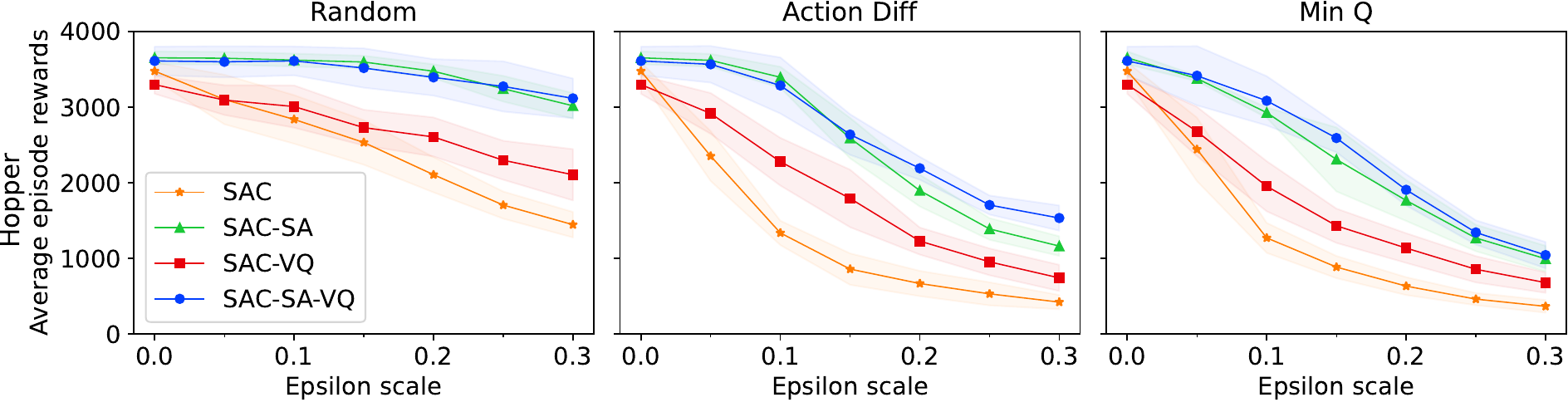}
    \includegraphics[width=0.5\textwidth]{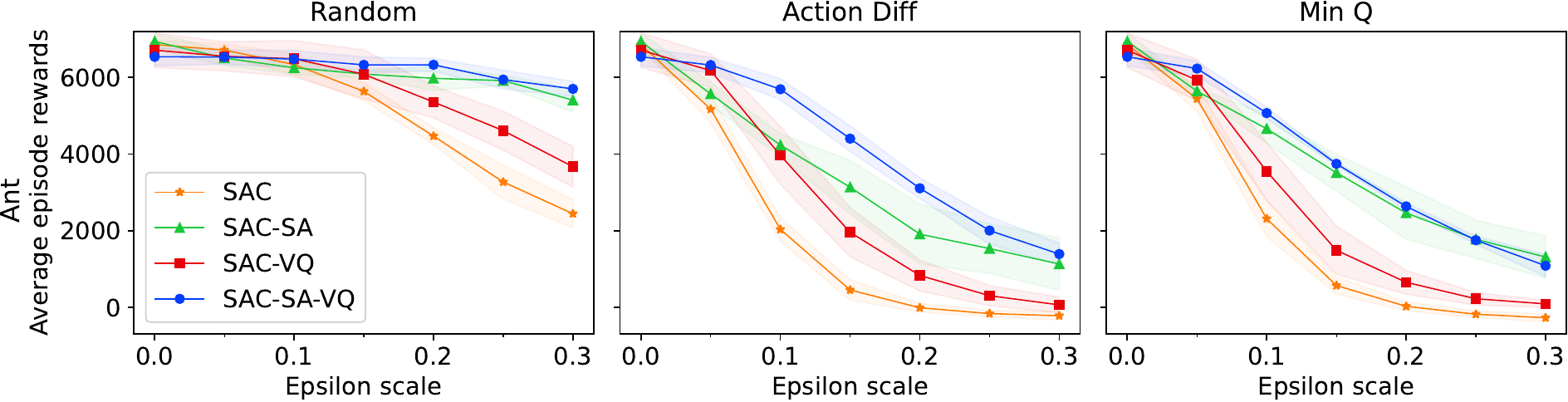}
    \includegraphics[width=0.5\textwidth]{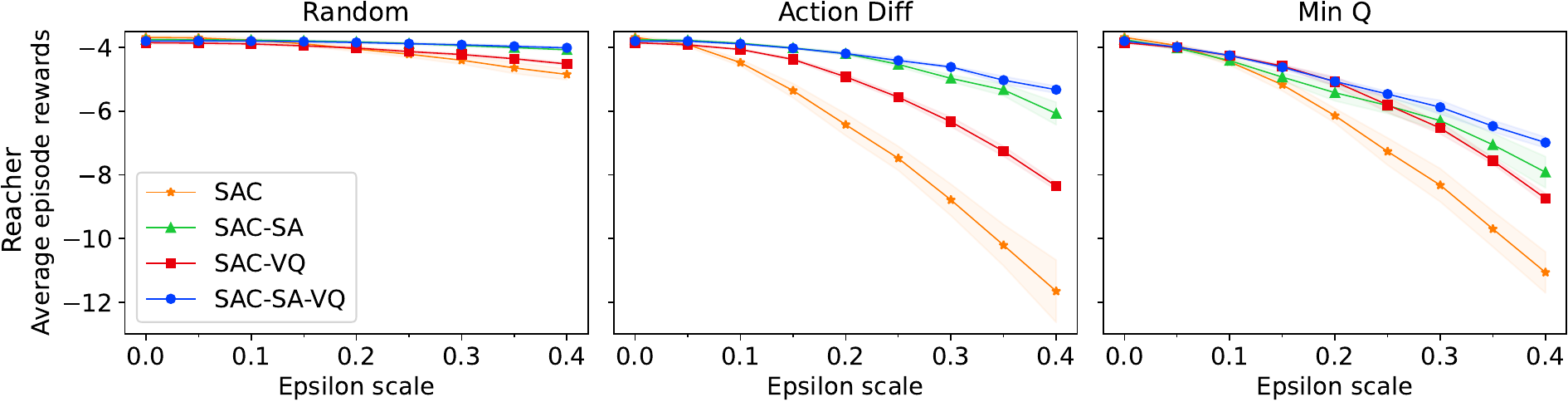}
    \includegraphics[width=0.5\textwidth]{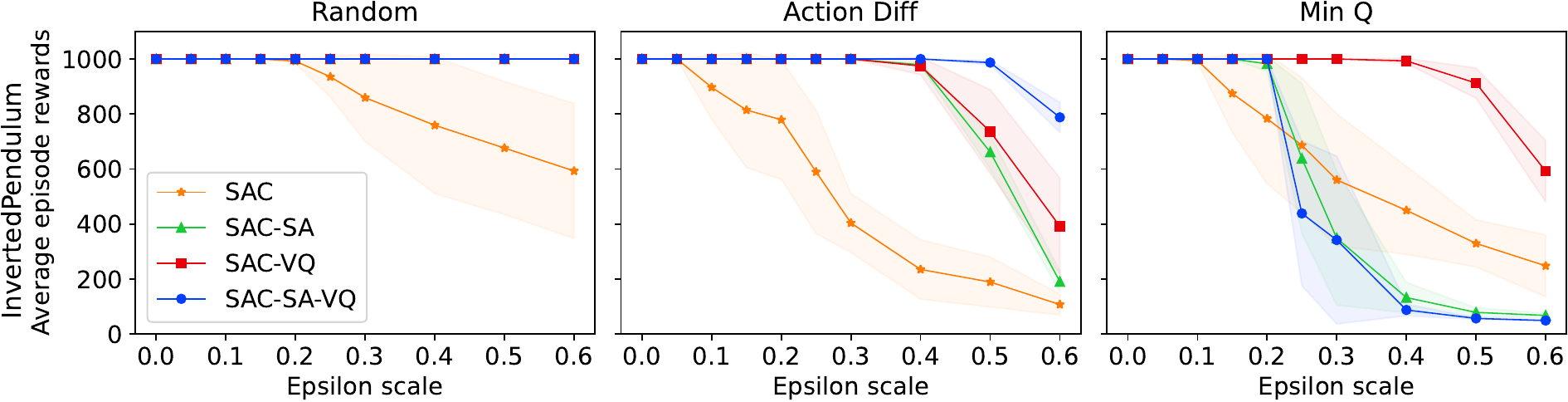}
    \caption{Comparisons under different attacks w.r.t. different scales of $\epsilon$ in online RL setting. The $y$-axis indicates the unnormalized return. The curve is averaged over five seeds, with $\pm1$ standard deviation shading.}
    \label{fig:online_aupc}
    \vskip -0.3in
\end{figure}

\begin{figure}[ht]
\centering
    \includegraphics[width=0.5\textwidth]{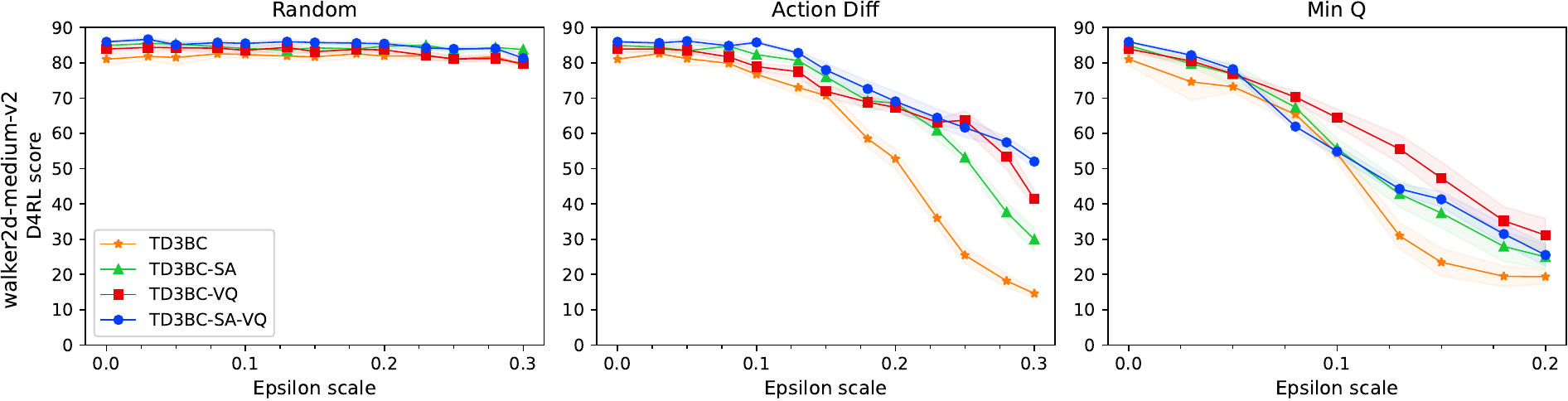}
    \includegraphics[width=0.5\textwidth]{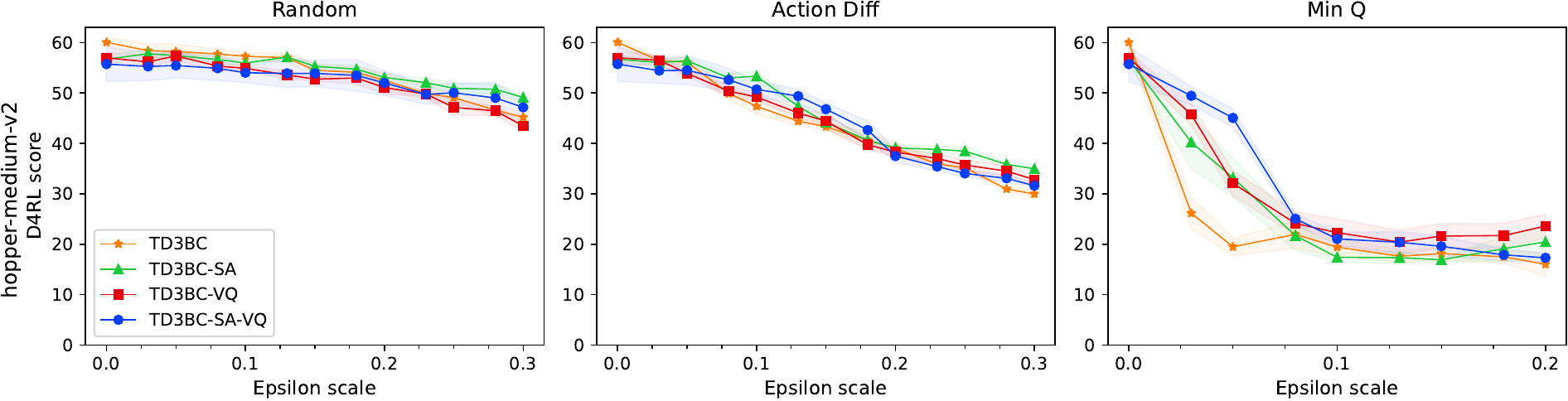}
    \includegraphics[width=0.5\textwidth]{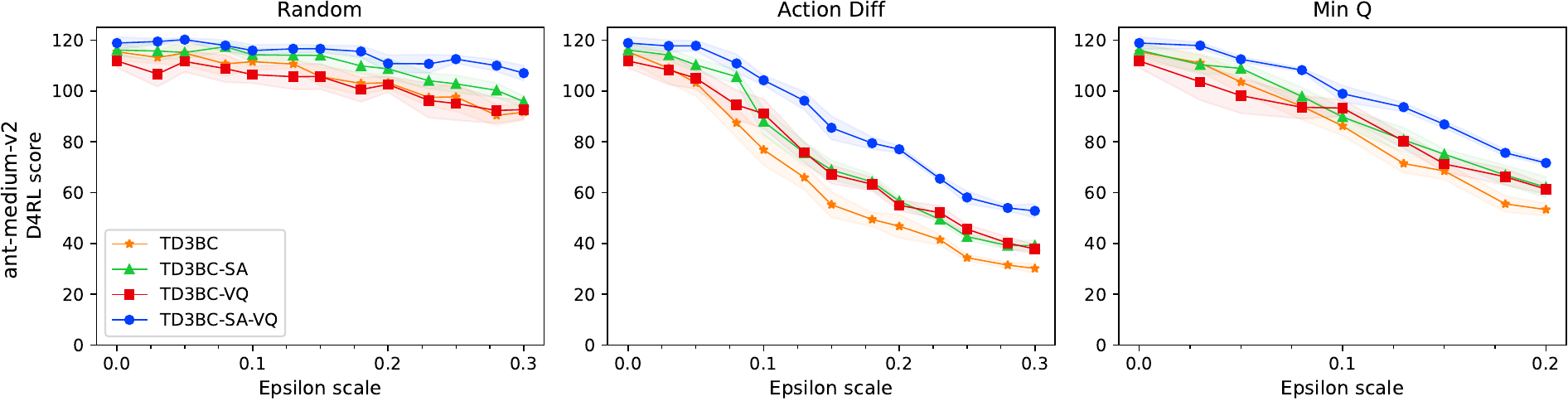}
    \caption{Comparisons under different attacks w.r.t. different budget $\epsilon$’s in offline RL setting. The $y$-axis indicates normalized return. The curve is averaged over five seeds, with $\pm1$ standard deviation shading.}
    \label{fig:offline_aupc}
    \vskip -0.3in
\end{figure}

\noindent
\textbf{Robustness Evaluation Under Different Attack Scales.} We provide the raw perturbation curves for online and offline RL setting in Fig. \ref{fig:online_aupc} and \ref{fig:offline_aupc}, respectively. These curves are used to obtain \textit{robustness score} in Tab. \ref{tab:aupc_online} and \ref{tab:aupc_offline}

\addtolength{\textheight}{-12cm}   

\end{document}